\documentclass[letterpaper, 10 pt, conference]{ieeeconf}  

\IEEEoverridecommandlockouts                              

\usepackage[dvipsnames,svgnames]{xcolor}
\usepackage{booktabs}       
\usepackage{nicefrac}       
\usepackage{microtype}      
\usepackage{multirow}
\usepackage{colortbl}
\usepackage{newtxmath}
\usepackage{pdfpages}
\usepackage{dcolumn}
\newcolumntype{d}[1]{D{.}{.}{#1}}
\usepackage{longtable}
\usepackage{threeparttable}
\usepackage{threeparttablex}
\usepackage{caption}
\usepackage{subfigure}
\usepackage{wrapfig}
\usepackage{graphicx}
\usepackage{amsmath,amscd,amsbsy,amsfonts,latexsym,url,bm,bbm,dsfont}
\usepackage{algorithm}
\usepackage{algorithmic}
\usepackage[normalem]{ulem}
\usepackage[algo2e,ruled,vlined,boxed,linesnumbered]{algorithm2e}
\usepackage[square,numbers]{natbib}
\usepackage[colorlinks,linkcolor=red,urlcolor=black]{hyperref}
\newcommand{\bd}{\boldsymbol}

\newcommand{\highlight}[1]{{\textcolor{MediumSeaGreen}{#1}}}

\title{\LARGE \bf
Boosting Offline Reinforcement Learning for Autonomous Driving \\ with Hierarchical Latent Skills
}

\author{Zenan Li$^{1,2}$, Fan Nie$^{2,3}$, Qiao Sun$^2$, Fang Da$^{4}$, and Hang Zhao$^{\dagger,1,2}$
\thanks{*This work was supported by Shanghai Qi Zhi Institute and QCraft Inc. Zenan Li and Fan Nie contribute equally.}
\thanks{$^{1}$Tsinghua University, $^{2}$Shanghai Qi Zhi Institute, $^{3}$Shanghai Jiao Tong University, $^{4}$QCraft Inc, $^{\dagger}$Corresponding author.}%
\thanks{Emails in order: \href{mailto:li-zn23@mails.tsinghua.edu.cn}{\emph{li-zn23@mails.tsinghua.edu.cn}}, \href{mailto:youluo2001@sjtu.edu.cn}{\emph{youluo2001@sjtu.edu.cn}}, \href{mailto:alan.qiao.sun@gmail.com}{\emph{alan.qiao.sun@gmail.com}}, \href{mailto:fang@qcraft.ai}{\emph{fang@qcraft.ai}}, \href{mailto:hangzhao@mail.tsinghua.edu.cn}{\emph{hangzhao@mail.tsinghua.edu.cn}}.}%
}

\begin{document}

\maketitle
\thispagestyle{empty}
\pagestyle{empty}

\begin{abstract}

Learning-based vehicle planning is receiving increasing attention with the emergence of diverse driving simulators and large-scale driving datasets. While offline reinforcement learning (RL) is well suited for these safety-critical tasks, it still struggles to plan over extended periods. In this work, we present a skill-based framework that enhances offline RL to overcome the long-horizon vehicle planning challenge. Specifically, we design a variational autoencoder (VAE) to learn skills from offline demonstrations. To mitigate posterior collapse of common VAEs, we introduce a two-branch sequence encoder to capture both discrete options and continuous variations of the complex driving skills. The final policy treats learned skills as actions and can be trained by any off-the-shelf offline RL algorithms. This facilitates a shift in focus from per-step actions to temporally extended skills, thereby enabling long-term reasoning into the future. Extensive results on CARLA prove that our model consistently outperforms strong baselines at both training and new scenarios. Additional visualizations and experiments demonstrate the interpretability and transferability of extracted skills.

\end{abstract}

\section{Introduction}
Learning-based vehicle planning is increasingly gaining its popularity with the advent of driving simulators~\cite{carla,intersim} and large-scale offline datasets~\cite{nuplan,waymo}. Among them, the performance of Imitation Learning (IL) algorithms is limited by the quality of expert data~\cite{ilreduction} as well as the covariate shift issue~\cite{ilaggregation,ilreduction}. Therefore, offline reinforcement learning (RL)~\cite{offlinerl,bcql,morel} emerges as an effective approach to learning policies that break the constraints of datasets, while avoiding risky exploration in real driving environments.

Despite its great potential, the application of offline RL in autonomous driving is relatively limited~\cite{offlinead,umbrella,higorl}, with the long-horizon problem presenting as a major obstacle~\cite{higorl,drivingskill}. Specifically, the reward signals in long-horizon tasks are sparse and delayed~\cite{hindsight,sparse}, and the agent must learn from them to plan a reasonable sequence of actions to avoid compounding errors over extended periods.

\begin{figure}[tb!]
    \centering
    \includegraphics[width=\linewidth]{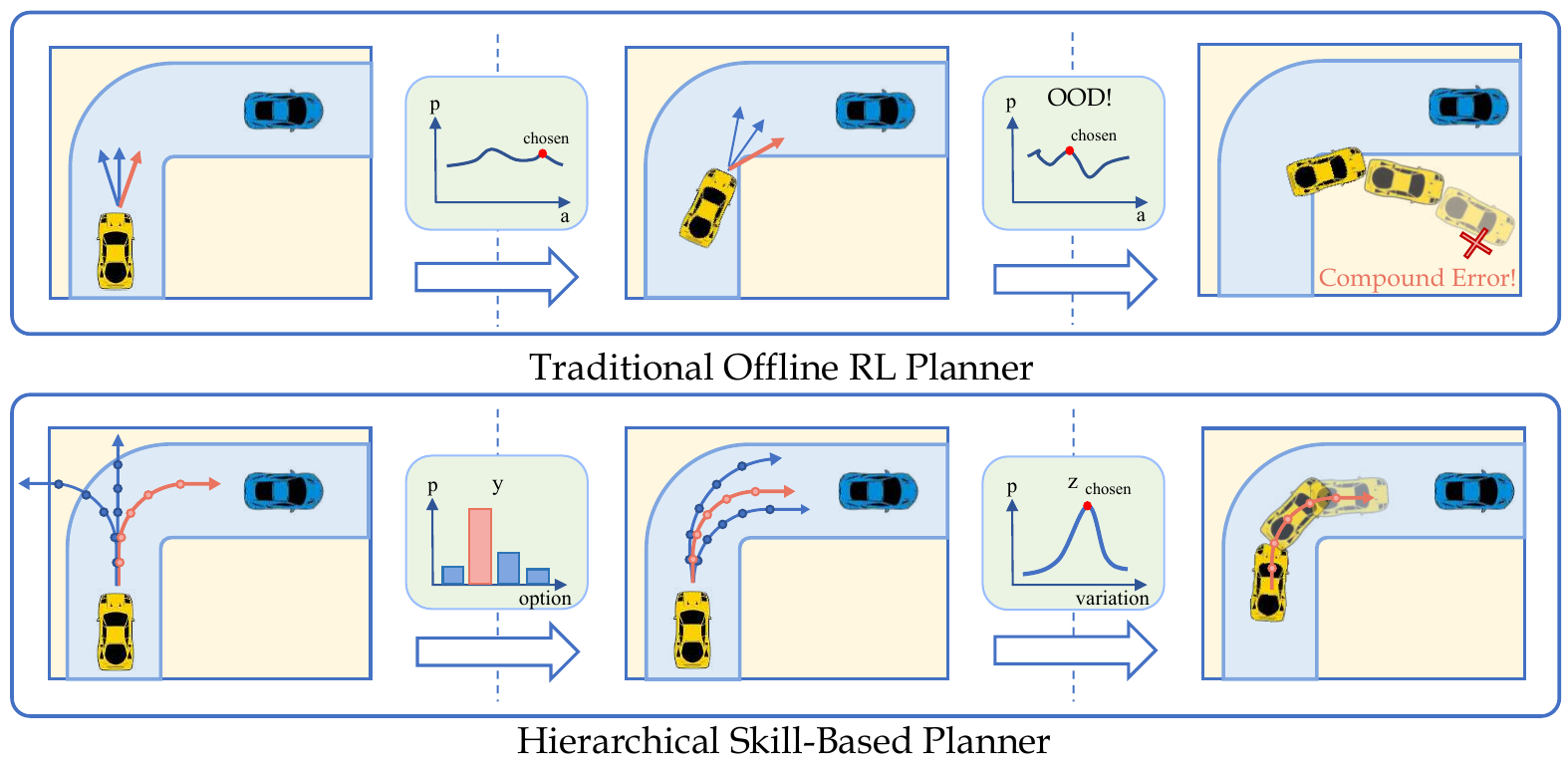}\vspace{-5pt}
    \captionsetup{font={footnotesize}}
    \caption{A motivating example. Top: Traditional offline RL planner samples action at each timestep, and is easy to go outside the distribution of training data, thus incurring compounding errors of actions. Bottom: Our proposed hierarchical skill-based planner first selects the discrete skill option (turning), and then decides the continuous variation for execution (radians), which can model a diverse set of driving skills. This multi-step skill execution enables long-term reasoning for the planner.}
    \label{fig:teaser}\vspace{-20pt}
\end{figure}

One powerful strategy to deal with long-horizon tasks is to adopt a hierarchical framework~\cite{hiil,higorl}, with the high-level policy producing a subgoal, and the low-level policy executing to approach the consecutive subgoals. As pointed out by~\cite{behavior} that human behaviors are fundamentally skill executions, our intuition is that the vehicle planning problem can also be solved by acquiring a set of \emph{driving skills} (e.g. cruising, nudging, and turning)~\cite{drivingskill,drivingskill1}. Therefore, we propose to interpret the subgoals as different driving skills to decompose the long-horizon task into a two-stage problem: (i) extract a set of reusable driving skills, and (ii) learn a skill-based policy. Essentially, skills provide a way to represent a sequence of actions. This temporal abstraction~\cite{temporalabs} allows the vehicle to reason and plan at a higher level of granularity~\cite{rethinking,skillprior}, taking longer-term consequences (so densely shaping reward signals) into consideration and improving long-horizon planning.

Several prior works have investigated techniques for extracting skills from offline datasets~\cite{skillprior,lmp,opal}, in general, they all employ a sequence-to-sequence conditional variational autoencoder (seq-to-seq CVAE)~\cite{gru,variational} to model the generative process of transition sequences, and extract a continuous latent space of skills describing different driving behaviors. This straightforward method can however suffer in the vehicle planning setting since the expert demonstrations are seriously \emph{unbalanced} in distribution, with a majority dominated by simple straight or static sequences. While this can be addressed by pre-filtering the dataset, a more fundamental challenge stems from the \emph{complex and versatile} nature of driving skills. Specifically, drivers must react aptly to different maps and environmental vehicle configurations, which encompass an infinite number of combinations, e.g. turning behaviors exhibit significant variations across different maps and road conditions.
Typically, VAEs are widely known to suffer from the posterior collapse problem~\cite{oversmoothvae,oversmoothvae2} under high input complexity, like that in Fig.~\ref{fig:opal_tsne}, where the encoder fails to model and distinguish between different driving skills.

The key to overcoming posterior collapse is to use more expressive latent distributions that can better capture diverse aspects of inputs~\cite{nvae}. Regarding this objective, we note the common structure of driving skills in Fig.~\ref{fig:teaser}: Generally, drivers first select within discrete skill options (e.g. turning), then determine continuous variations (e.g. radians and speeds) during execution. Therefore, we suggest to inject this prior into the generative process and extract a hierarchical latent space instead of simple gaussian distributions. Specifically, our VAE encoder is two-branch: the discrete branch produces logits that are passed through a gumbel-softmax activation~\cite{gumbel} to get approximately one-hot variables; the continuous branch functions as an ensemble, with discrete outputs acting as gates to select a member network for computing the final skill variable. In this sense, utilizing the discrete branch to distinguish between different skill options, the member networks in the continuous ensemble just need to fit variations within a specific skill option, which is a relatively simple distribution, thereby reducing the risk of posterior collapse. 

After distilling skills from data, any off-the-shelf offline RL algorithm~\cite{bcql,cql,iql} can be used to learn the high-level skill-based policy using relabeled transition data. The VAE decoder is employed as the low-level action decoder, which can also be finetuned with conditional behavior cloning (BC).

To summarize, we make the following contributions:
\begin{itemize}
\item We present \textbf{\underline H}ierarchical \textbf{\underline s}kill-based \textbf{\underline O}ffline Reinforcement Learning for \textbf{\underline V}ehicle \textbf{\underline P}lanning (HsO-VP), a framework that boosts offline RL to solve the challenging long-horizon vehicle planning task.
\item We propose a two-branch network architecture for the VAE encoder, thus forming a hierarchical latent space that can effectively address the posterior collapse issue during driving skill extraction.
\item We comprehensively evaluate HsO-VP on CARLA~\cite{carla}, where it achieves consistent performance improvements over the baselines (4.0\% at training scenarios and 6.4\% at new scenarios). Besides, the learned skills are also verified to possess interpretability and transferability.
\end{itemize}

\section{Related Works}
The objective of vehicle planning is to efficiently drive the ego-vehicle to the destination while conforming to some safety and comfort constraints, which is essentially a sequential decision problem. Recently, learning-based planning algorithms~\cite{chauffeurnet,nsm,roach,mbil} are progressively gaining popularity over conventional rule-based algorithms~\cite{optimize,cilqr} for their better scalability to different driving scenarios. In this section, we review works about offline RL and skill-based algorithms as the foundation of our work and highlight the differences and contributions of HsO-VP.

\textbf{Offline RL:} Offline RL algorithms learn policies from a static dataset without interacting with the real environment~\cite{offlinerl}, which is particularly useful in scenarios where interaction can be prohibitively expensive or risky like autonomous driving. Unlike IL, offline RL algorithms have been proven to be capable of surpassing the performance of experts~\cite{bcql,cql,iql}. Generally, they utilize policy regularization~\cite{bcql,minimal} and out-of-distribution (OOD) penalization~\cite{cql,uncertainty} to avoid value overestimation.
In this paper, we pioneer to deploy offline RL algorithms to the long-horizon vehicle planning task, where the reward may be sparse and the extrapolation error will be accumulated~\cite{hindsight,sparse}. Although there have been several works~\cite{offlinead,umbrella,higorl} attempting to apply offline RL to vehicle planning, few of them~\cite{higorl} directly deal with the long-horizon challenge. As the most relevant work to our paper, HiGoC~\cite{higorl} employs a hierarchical planning framework (not open-sourced so we choose IRIS~\cite{iris} instead as the comparing baseline), where the high-level policy produces subgoal states through optimization, and the low-level policy is trained by offline RL conditioned on subgoals. In contrast, HsO-VP utilizes skill completion as subgoals instead of imagined target states, which offers better interpretability in driving tasks. Besides, offline RL is conducted on the high-level skill-based policy in HsO-VP, which enables long-term reasoning during training. 

\textbf{Skill-based algorithms:} Skills, also known as primitives or macro actions, are widely used in RL to enable knowledge transfer on new tasks~\cite{skillprior,skillmetarl} or facilitate long-horizon training~\cite{opal,tacorl}. Previous works~\cite{drivingskill1,drivingskill2} have attempted to manually design driving skills for specific tasks. While these task-specific skill designs can succeed in certain well-defined tasks such as overtaking a car, they limit the flexibility and expressiveness in complex driving scenarios~\cite{drivingskill}. In addition to this, another line of research explores skills from the perspective of unsupervised segmentation of reusable behaviors in temporal data~\cite{skillprior,lmp,opal,skillmetarl,tacorl}. Generally, they use a VAE to extract continuous skill representations from expert transition sequences. Once the skills have been extracted, a new high-level skill-based policy is learned from relabeled transitions by online~\cite{skillprior,lmp,skillmetarl} or offline RL~\cite{opal,tacorl}, with the VAE decoder acting as the low-level policy that outputs per-step actions conditioned on high-level skills.
In this paper, we find vanilla VAEs not expressive enough for modeling the versatile driving demonstrations. As a treatment, we introduce another categorical latent variable to form a hierarchical latent space that can capture both discrete options and continuous variations of driving skills, enhancing the interpretability and transferability of learned skills. 
Different from~\cite{drivingskill} that learns driving skills from sampled trajectories, we extract interpretable skills from expert demonstrations and focus on the offline RL setting.

\section{Preliminary}\label{sec:pre}
\begin{figure*}[tb!]
    \centering
    \includegraphics[width=0.9\linewidth]{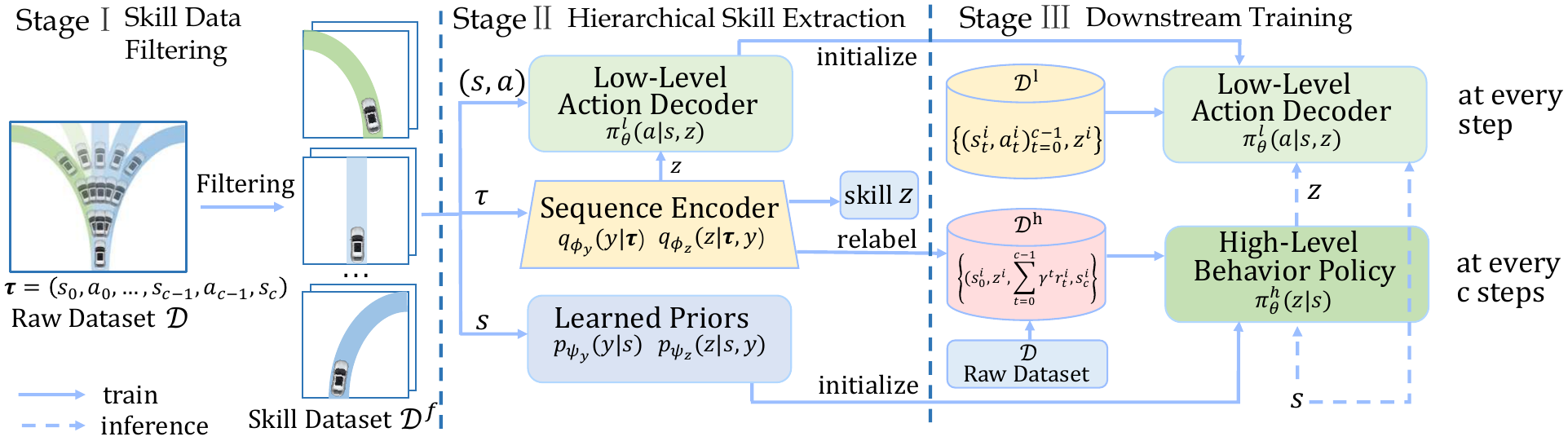}\vspace{-5pt}
    \captionsetup{font={footnotesize}}
    \caption{Overview of HsO-VP. Left: Pre-filter the raw dataset to get the balanced skill dataset. Middle: Main components of our proposed VAE with hierarchical latent space for driving skill extraction. Right: Downstream training for our skill-based policy. Both high and low-level policies are initialized from Stage II, and trained by offline RL and conditional BC from the relabeled datasets, respectively.}
    \label{fig:framework}\vspace{-20pt}
\end{figure*}
We introduce preliminary knowledge for HsO-VP in this section. To keep notations concise, we use subscripts $t, c$ or numbers for variables at specific timesteps, Greek letter subscripts for parameterized variables, and bold symbols to denote variables spanning multiple timesteps.

We consider learning in a Markov decision process (MDP)~\cite{mdp} denoted by the tuple $\mathcal{M}=(\mathcal{S},\mathcal{A},\mathcal{P},r,\gamma)$, where $\mathcal{S}$ and $\mathcal{A}$ are the state space and the action space, respectively. Given states $s,s'\in\mathcal{S}$ and action $a\in\mathcal{A}$, $\mathcal{P}(s'|s,a):\mathcal{S}\times\mathcal{A}\times\mathcal{S}\rightarrow[0,1]$ is the transition probability distribution and $r(s,a):\mathcal{S}\times\mathcal{A}\rightarrow\mathbb R$ defines the reward function. Besides, $\gamma\in(0,1]$ is the discount factor. The agent takes action $a$ at state $s$ according to its policy $\pi(a|s):\mathcal{S}\times\mathcal{A}\rightarrow[0,1]$. The goal of \emph{online RL} is to find a policy $\pi$ that maximizes the total expected return: $J=\mathbb{E}_{a_t\sim\pi(\cdot|s_t),s_{t+1}\sim \mathcal{P}(\cdot|s_t,a_t)}\big[\sum_{t=0}^{T-1}\gamma^{t}r(s_t,a_t)\big]$
by learning from the transitions $(s,a,r,s')$ through interacting with the environment in an online manner.
\emph{Offline RL,} instead makes use of a static dataset with $N$ trajectories $\mathcal{D}=\{\bd{\tau}_i\}_{i=1}^{N}$ collected by certain behavior policy $\pi_b$ to learn a policy $\pi$ that maximizes $J$, thereby avoiding safety issues during online interaction. Here $\bd{\tau}=\big\{(s_t,a_t,r_t,s_t')\big\}_{t=0}^{T-1}$ is a collected interaction trajectory composed of transitions with horizon $T$.

\section{Approach: HsO-VP}\label{sec:approach}
In this section, we begin with an overview of HsO-VP's insight and architecture, followed by detailed explanations of composition modules used in the approach.

\vspace{-5pt}\subsection{Model Overview}\label{sec:overview}
An overview of the proposed approach HsO-VP is illustrated in Fig.~\ref{fig:framework}. Specifically, it consists of three stages: First, we conduct \emph{skill data filtering} for a balanced skill dataset, facilitating subsequent VAE training. Second, a two-branch VAE is specially designed for \emph{hierarchical skill extraction}, modeling structured and complex skill representations to overcome the posterior collapse problem. Finally, the extracted skills, which function as temporal abstractions to enable long-horizon planning, are used to relabel the dataset for \emph{downstream training}. In the following, we introduce each stage with more details. For convenience, we use the terms `skill' and `behavior' interchangeably.

\vspace{-5pt}\subsection{Skill Data Filtering}\label{sec:filter}
The temporally extended skills can be a solution to boost offline RL for long-horizon vehicle planning, which enables the vehicle to plan at a higher level of granularity, thereby facilitating long-term reasoning into the future~\cite{rethinking, skillprior}. However, raw expert demonstrations are not well-suited for skill extraction due to their unbalanced distribution (with straight and static trajectories occupying the majority). Therefore, we propose to pre-filter the transition sequences $\bd{\tau}=(s_0,a_0,...,s_{c-1},a_{c-1},s_c)$, where $c$ stands for the skill length, for a balanced filtered skill dataset $\mathcal D^f$. As an auxiliary preprocessing stage, meticulous data management is not necessary. Given this and since it's not the primary focus of this work, here we briefly introduce our filtering principle.

Since the action sequences can already reflect the driving skills to a great extent (e.g. the turning and straight behaviors can be easily distinguished by corresponding action sequences), we concatenate these $c$-step low-dimensional actions into vectors $\bd{a}_{0:c-1}$ and conduct clustering~\cite{kmeans}. With different clusters coarsely representing different skills, our objective is then to maintain uniformity in the sequence count within each cluster. In practical implementation, we control and balance sequence counts through a set of thresholds. Specifically, we choose a threshold for each cluster, filter and retain action sequences whose pairwise distances are greater than (different therefore representative) the threshold.

\vspace{-5pt}\subsection{Hierarchical Skill Extraction}
Now with the filtered balanced skill dataset, we can step to extract reusable driving skills through variational inference (VI)~\cite{variational}, similar to previous works~\cite{skillprior, lmp,skillmetarl}.

\textbf{Generative Process:} The key of VI is to model the generative process of transition sequences and their corresponding skill variables $z$. Generally, it can be written as follows:
\begin{equation}\label{eq:genproc_ex}
\begin{split}
    p(\bd{\tau},{z})=&p(\bd{s}_{0:c},\bd{a}_{0:c-1},{z})=p({s}_0)p({z}|{s}_0)\cdot\\ &\prod_{t=0}^{c-1}p({a}_{t}|{s}_{t},{z})p({s}_{t+1}|{s}_{t},{a}_{t}).
\end{split}
\end{equation}
In consideration of vehicle planning, however, the skills can be more complex and versatile than those in other fields~\cite{drivingskill,drivingskill1} across different maps and road conditions. This complexity can give rise to posterior collapse~\cite{oversmoothvae,oversmoothvae2}, where the VAE fails to capture complex dependencies of input data and results in a degenerated indistinguishable latent space. So to boost the expressive capability of modeling driving skills, we propose to explicitly model the skill structure for better learning the latent representations. Noticing that driving behavior is usually conducted by first selecting a discrete skill option (e.g. turning) and then determining the specific variations (e.g. speed) to be executed, we introduce another categorical variable ${y}\in\mathds{1}^K$ to capture discrete options and form a hierarchical latent space:
\begin{equation}\label{eq:genproc}
\begin{split}
    p(\bd{\tau},{y},{z})=&p(\bd{s}_{0:c},\bd{a}_{0:c-1},{y},{z})=p({s}_0)p({y}|{s}_0)\cdot\\ &p({z}|{y},{s}_0)\prod_{t=0}^{c-1}p({a}_{t}|{s}_{t},{z})p({s}_{t+1}|{s}_{t},{a}_{t}).
\end{split}
\end{equation}
Through this hierarchical decomposition, $z$ is required to capture variations within a specific discrete option (depending on $y$) instead of the full data distribution, which is relatively simpler to model, thereby mitigating the risk of posterior collapse. Practically, we model the generative priors by some flexible parametric distributions $p_{\psi_y}(y|s_0)p_{\psi_z}(z|y,s_0)$. To keep notations consistent with the RL setting, we use a low-level action decoder $\pi_\theta^l(a_t|s_t,z)$ to substitute for $p(a_t|s_t,z)$. Besides, the state transition distribution $p(s_{t+1}|s_t,a_t)$ is fixed and decided by the environment.

\textbf{Learning Objective:} To infer latent skills from data, we introduce transition sequence encoders $q_{\phi_y}(y|\bd{\tau})q_{\phi_z}(z|\bd{\tau},y)$ to approximate the intractable posterior $p(y|\bd{\tau})p(z|\bd{\tau},y)$. Based on these, the target of VI is to maximize the \emph{conditional probability} $\log p_{\theta,\psi}(\bd{\tau}|s_0)$ of the observed transition sequence $\bd{\tau}$ w.r.t. the initial state $s_0$, which is deduced to be larger than the Evidence Lower BOund (ELBO)~\cite{variational}. Our learning objective is therefore transformed to maximize the ELBO:
\begin{align}
    \max_{\phi,\psi,\theta}&\mathbb E_{\bd{\tau}\sim\mathcal D^f,{y}\sim q_{\phi_y}({y}|\bd{\tau}),{z}\sim q_{\phi_z}({z}|\bd{\tau},{y})}\bigg[\underbrace{\sum_{t=0}^{c-1}\log\pi_\theta^l(a_t|s_t,{z})}_{\text{reconstruction loss}\ \mathcal L_{\text{recon}}}\bigg] \nonumber \\
    -\beta_z&\mathbb E_{\bd{\tau}\sim\mathcal D^f,{y}\sim q_{\phi_y}({y}|\bd{\tau})}\bigg[\underbrace{D_{\text{KL}}\Big(q_{\phi_z}({z}|\bd{\tau},{y})||p_{\psi_z}({z}|s_0,{y})\Big)}_{\text{continuous regularization}\ \mathcal L_{\text{KL}}^{\text{cont}}}\bigg] \nonumber \\
    -\beta_y&\mathbb E_{\bd{\tau}\sim\mathcal D^f}\bigg[\underbrace{D_{\text{KL}}\Big(q_{\phi_y}({y}|\bd{\tau})||p_{\psi_y}({y}|s_0)\Big)}_{\text{discrete regularization}\ \mathcal L_{\text{KL}}^{\text{dis}}}\bigg]. 
\end{align}
For practical implementation, we adopt the $\beta$-VAE~\cite{betavae} trick and weigh the KL terms with $\beta_y$ and $\beta_z$, respectively.

\begin{figure}[tb!]
    \centering
    \includegraphics[width=0.8\linewidth]{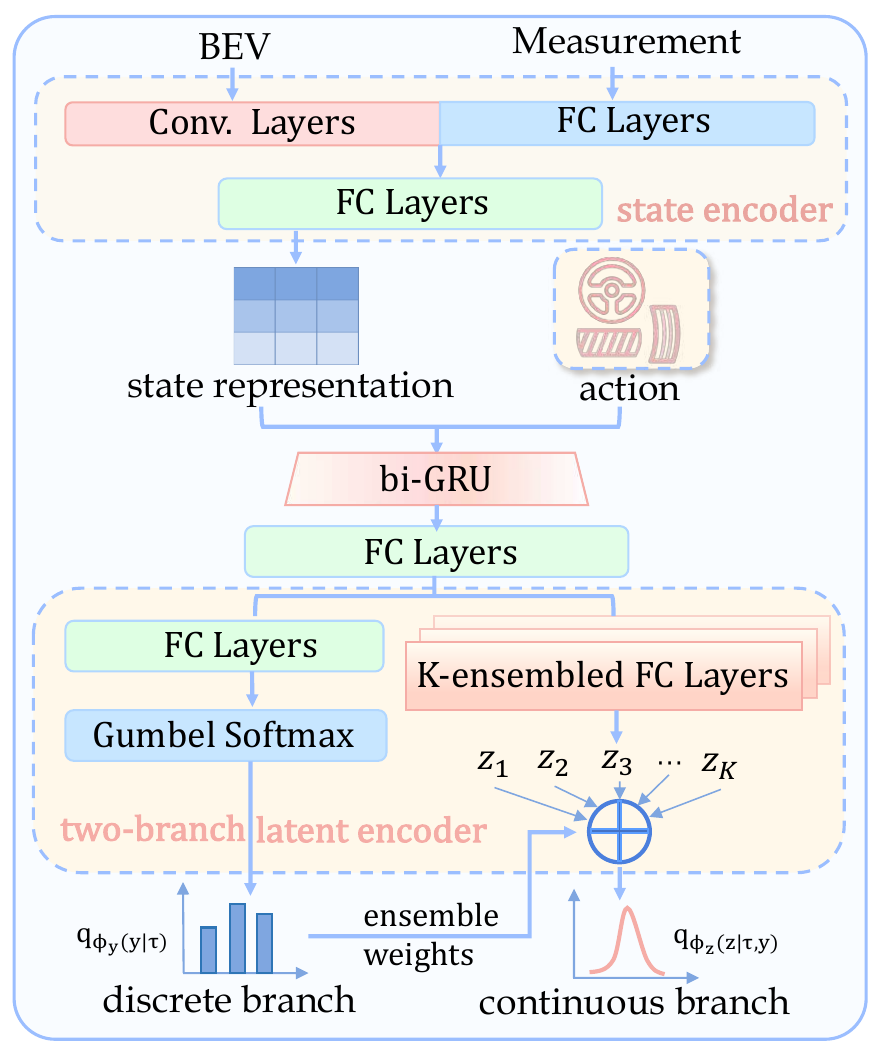}\vspace{-5pt}
    \captionsetup{font={footnotesize}}
    \caption{The two-branch network architecture of our proposed transition sequence encoder. Note that the input to bi-GRU is actually composed of states and actions over several timesteps.}
    \label{fig:net}\vspace{-20pt}
\end{figure}

\textbf{Network Instantiation:} Finally, we specify the network architecture of the proposed modules for skill extraction. For all vehicle planners implemented in this paper, we use the same state input composed of a bird's-eye view (BEV) semantic segmentation image and a measurement vector.
\begin{itemize}
\item $\bm{[}\text{Sequence encoders}\bm{]}$ $q_{\phi_y}(y|\bd{\tau})$ and $q_{\phi_z}(z|\bd{\tau},y)$: The architecture of our proposed sequence encoder is shown in Fig.~\ref{fig:net}. For the state representation, we use convolutional layers to encode the BEV and fully connected (FC) layers for the measurement vector. Afterward, a gated recurrent unit (GRU)~\cite{gru} is adopted to process the $c$-step transition inputs. For modeling the driving skill structure, we use a two-branch architecture after the GRU: the discrete branch takes in the GRU outputs and produces $K$-dimensional logits for $y$; the continuous branch is composed of $K$-ensembled FC layers, which outputs final continuous skill variable $z$ by weighted summation with $y$ acting as coefficients. Notably, we use gumbel-softmax~\cite{gumbel} activation to sample approximately one-hot discrete variables $y$, acting as a gate to guide members in the ensemble to learn about skills from different options.
\item $\bm{[}\text{Low-level action decoder}\bm{]}$ $\pi_\theta^l(a|s,z)$: We use a GRU decoder that takes in state and skill embeddings to output low-level executable actions. 
\end{itemize}
The learnable priors $p_{\psi_y}(y|s)$ and $p_{\psi_z}(z|s,y)$ and the high-level behavior policy $\pi_\theta^h(z|s)$ just share the same architecture as sequence encoders except that we remove the GRU and directly use the single-step state representation for $y$ and $z$.

\subsection{Downstream Training}\label{sec:relabel}
Once the skills have been learned from $\mathcal{D}^f$ in terms of sequence encoders $q_{\phi_y}({y}|\bd{\tau})q_{\phi_z}({z}|\bd{\tau},{y})$, action decoder $\pi_\theta^l({a}|{s},{z})$, and priors $p_{\psi_y}({y}|{s})p_{\psi_z}({z}|{s},{y})$, HsO-VP can then apply these modules to learn a skill-based vehicle planner.

\textbf{Relabeled skill dataset:} To learn a high-level skill-based policy $\pi_\theta^h({z}|{s})$ with offline RL, we need to relabel transitions in $\mathcal{D}$ (w/o filtering). Specifically, for each $c$-step sequence sampled from the dataset $\bd{\tau}\sim\mathcal D$, we label its corresponding skill by sequence encoders $z\sim \mathbb E_{y\sim q_{\phi_y}({y}|\bd{\tau})}q_{\phi_z}({z}|\bd{\tau},{y})$, thereby creating a new dataset $\mathcal{D}^h=\{(s_0^i,z^i,\sum_{t=0}^{c-1}\gamma^tr_t^i,s_c^i)\}_{i=1}^{N^h}$ that treats skills as actions. 

\textbf{Downstream Learning Procedure:} Given the relabeled dataset $\mathcal D^h$, any off-the-shelf offline RL algorithm can be used to learn $\pi_\theta^h(z|s)$. In this paper, we choose IQL~\cite{iql}, which is one of the state-of-the-art (SOTA) offline RL algorithms for end-to-end policy training. 
Besides, the high-level behavior policy is initialized with the learned priors $\mathbb E_{y\sim p_{\psi_y}(y|s)}p_{\psi_z}(z|s,y)$, which can be a good starting point to solve some short-horizon tasks. 

During inference, $\pi_\theta^h(z|s)$ is invoked at every $c$-steps, while $\pi_\theta^l(a|s,z)$ is used to give executable low-level actions for the vehicle. To ensure that the $c$-step transitions $(s_t,a_t)_{t=0}^{c-1}$ remain consistent with the labeled latent skill $z$, we can also optionally finetune $\pi_\theta^l(a|s,z)$ on $\mathcal D^l=\{(s_t^i,a_t^i)_{t=0}^{c-1},z^i\}_{i=1}^{N^l}$ with a skill-conditioned behavior cloning (BC) loss:
\begin{equation}\label{eq:condbc}
    \min_\theta \mathbb E_{(\bd{\tau},z)\sim\mathcal D^l}\bigg[-\sum_{t=0}^{c-1}\log\pi_\theta^l(a_t|s_t,{z})\bigg].
\end{equation}

\section{Evaluation}
In this section, we do extensive experiments to answer the following questions. \highlight{\textbf{Q1:}} How effective is HsO-VP compared to baselines when deployed in seen or unseen driving scenarios? \highlight{\textbf{Q2:}} How do the composed modules impact HsO-VP's performance? \highlight{\textbf{Q3:}} Do skills extracted by HsO-VP possess interpretability and transferability?

\subsection{Experiment Setup}\label{sec:expsetup}

\textbf{Datasets:} The training dataset $\mathcal {D}$ is collected in the CARLA simulator with an expert RL-based planner~\cite{roach} trained with BEV semantic images and privileged vehicle measurements. More specifically, we collect data at 10Hz from 4 different training towns (Town01, Town03, Town04, Town06) and 4 weather conditions (ClearNoon, WetNoon, HardRainNoon, ClearSunset) for a total of 10 hours of driving data. At each timestep, we save a tuple $(i_t,m_t,a_t,r_t)$, with $i_t\in\mathbb R^{15\times 192\times 192}$ the BEV semantic image, $m_t\in\mathbb R$ the velocity of the ego-vehicle, $a_t\in[-1,1]^2$ the action for steering and acceleration executed by the expert, and $r_t\in\mathbb R$ the current step reward from the environment. Our reward design is modified from~\cite{roach}, decided by elements including speed, safety, comfort, etc. Besides, to study skill transferability, we also collect 5 hours of driving data from Town05.  

\textbf{Metrics:} We report metrics from the CARLA challenge~\cite{carla,leaderboard} to measure planners' driving performance: infraction score, route completion, success rate, and driving score. Besides, as done in~\cite{mbil}, we also report the normalized reward (the ratio of total return to the number of timesteps) to reflect the driving performance at timestep level. Among them, driving score is the most significant metric for evaluating planners' performance, which is a weighted score of various indicators like driving efficiency, safety, and comfort.

\textbf{Baselines:} First, we choose two IL baselines: vanilla Behavior Cloning (BC) and Monotonic Advantage Re-Weighted Imitation Learning (MARWIL)~\cite{marwil}. Apart from IL baselines, we also include SOTA offline RL baselines: Conservative Q-Learning (CQL)~\cite{cql}, Implict Q-Learning (IQL)~\cite{iql}. Finally, we add two SOTA hierarchical offline RL algorithms: Implicit Reinforcement without Interaction at Scale (IRIS)~\cite{iris} that generates subgoals using offline RL, and Offline Primitives for Accelerating offline reinforcement Learning (OPAL)~\cite{opal} which extracts skills using vanilla VAEs, as rigorous baselines.

\textbf{Implementation Details:} Training is conducted on 8 NVIDIA 3090 GPUs, while inference is performed on one NVIDIA 3090 GPU to ensure a fair comparison. For skill filtering and extraction, we remove background vehicles to prevent any detrimental impact on the extraction of the ego-vehicle's skills due to their stochastic movements~\cite{drivingskill}. These vehicles are retained in the downstream offline RL process for the ego-vehicle to make reasonable reactions. For filtering, the cluster number is set as 6 and we retain approximately 50\% of expert data. The skill length $c$ is set as 10 (1 second of driving). The skill extraction networks are trained from scratch using the Adam optimizer (with a learning rate of 0.0001) for 1000 epochs, employing a batch size of 64. The gumbel temperature is 0.1, regularization factors $\beta_y$ and $\beta_z$ are set to 0.01, and the number of discrete skill options $K$ (i.e. dimension of $y$) is set as 6. The downstream training process shares the same optimizer configuration with the skill extraction stage, and the discounted factor $\gamma$ is set to 0.99. More details will be public in codes when published.

\vspace{10pt}\subsection{Driving Performance}\label{sec:drive_exp}
\begin{table}[tb!]
    \centering
    \captionsetup{font={footnotesize}}
    \caption{Driving performance on a train town and train weather conditions in CARLA. Mean and standard deviation are computed over 3 evaluation seeds. See the main text for detailed meanings of HsO-VP variants. All metrics are recorded in percentages (\%) except the normalized reward. The best results are in bold and our method is colored in gray.}\vspace{-5pt}
    \label{tab:tt_carla}
    \resizebox{\linewidth}{!}{
    \begin{tabular}{l|ccccc}
    \toprule[1.0pt]
        \multirow{2}{*}{Planner} & Driving & Success & Route & Infraction & Norm. \\ 
        & Score$\uparrow$ & Rate$\uparrow$ &Co.$\uparrow$ & Score$\uparrow$ & Reward$\uparrow$\\ \midrule[0.6pt]
        BC & $65.7\pm2.8$ & $44.2\pm6.2$ & $78.3\pm3.1$ & $73.3\pm1.8$ &$0.40\pm0.08$\\
        MARWIL~\cite{marwil} & $66.8\pm2.3$ & $45.0\pm7.1$ & $80.7\pm1.2$ & $73.4\pm1.2$ &$0.46\pm0.01$\\
        CQL~\cite{cql} & $69.4\pm3.6$ & $50.3\pm8.5$ & $81.5\pm3.5$ & $76.2\pm3.0$ & $0.56\pm0.06$ \\
        IQL~\cite{iql} & $70.7\pm4.4$ & $51.2\pm6.6$ & $82.0\pm2.0$ & $76.7\pm5.0$ & $0.57\pm0.05$ \\
        IRIS~\cite{iris} & $72.8\pm2.8$ & $54.0\pm4.8$ & $85.1\pm2.6$ & $78.3\pm3.2$ & $0.60\pm0.06$\\
        OPAL~\cite{opal} & $72.1\pm2.4$ & $53.2\pm5.5$ & $84.3\pm4.1$ & $77.9\pm2.4$ & $0.59\pm0.08$ \\ \midrule[0.6pt]
        HsO-VP Raw & $68.3\pm2.4$ & $47.3\pm6.6$ & $81.5\pm3.5$ & $74.9\pm1.4$ & $0.51\pm0.04$ \\
        HsO-VP w/o BC & $74.1\pm1.2$ & $57.8\pm4.4$ & $83.3\pm2.1$ & $80.2\pm2.2$ & $0.62\pm0.03$ \\ \rowcolor{gray!40}
        HsO-VP & \bd{$76.8\pm1.4$} & \bd{$59.2\pm4.5$} & \bd{$86.8\pm3.3$} & \bd{$82.4\pm1.4$} & \bd{$0.65\pm0.03$} \\ \midrule[0.6pt]
        Expert~\cite{roach} & $78.7\pm1.1$ & $60.0\pm2.0$ & $89.5\pm5.4$ & $84.1\pm1.2$ & $0.69\pm0.03$ \\
    \bottomrule[1.0pt]
    \end{tabular}}\vspace{-10pt}
\end{table}
\begin{table}[tb!]
    \centering
    \captionsetup{font={footnotesize}}
    \caption{Driving performance on a new town and new weather conditions in CARLA. Transfer learning results are colored in green. The best results in bold do not take into account the transfer learning results.}\vspace{-5pt}
    \label{tab:nn_carla}
    \resizebox{\linewidth}{!}{
    \begin{tabular}{l|ccccc}
    \toprule[1.0pt]
        \multirow{2}{*}{Planner} & Driving & Success & Route & Infraction & Norm. \\ 
        & Score$\uparrow$ & Rate$\uparrow$ &Co.$\uparrow$ & Score$\uparrow$ & Reward$\uparrow$\\ \midrule[0.6pt]
        BC & $68.8\pm2.5$ & $48.3\pm6.2$ & $80.0\pm7.1$ & $74.7\pm0.8$ & $0.35\pm0.17$\\
        MARWIL~\cite{marwil} & $70.9\pm6.7$ & $58.1\pm9.5$ & $83.3\pm2.4$ & $76.0\pm6.6$ & $0.41\pm0.13$\\
        CQL~\cite{cql} & $72.8\pm1.6$ & $61.3\pm6.2$ & $85.1\pm3.1$ & $77.1\pm3.0$ & $0.54\pm0.01$ \\
        IQL~\cite{iql} & $73.4\pm3.8$ & $63.0\pm4.1$ & $86.3\pm3.1$ & $77.8\pm4.8$ & $0.56\pm0.03$ \\
        IRIS~\cite{iris} & $76.4\pm3.3$ & $65.8\pm5.2$ & $91.1\pm2.7$ & $79.4\pm3.2$ & $0.61\pm0.02$\\
        OPAL~\cite{opal} & $74.3\pm3.4$ & $64.2\pm6.5$ & $88.2\pm4.4$ & $78.8\pm2.8$ & $0.58\pm0.04$ \\ \rowcolor{green!40}
        OPAL Tran. & $75.0\pm2.2$ & $64.9\pm4.5$ & $89.4\pm3.3$ & $79.2\pm3.4$ & $0.60\pm0.05$ \\ \midrule[0.6pt]
        HsO-VP Raw & $75.7\pm4.8$ & $60.8\pm2.8$ & $83.3\pm3.8$ & $77.8\pm3.3$ & $0.55\pm0.08$ \\
        HsO-VP w/o BC & $80.1\pm2.3$ & $70.7\pm1.5$ & $90.5\pm1.2$ & $81.6\pm2.6$ & $0.63\pm0.03$ \\ \rowcolor{gray!40}
        HsO-VP & \bd{$82.8\pm1.8$} & \bd{$71.7\pm2.5$} & \bd{$94.9\pm2.8$} & \bd{$83.4\pm2.9$} & \bd{$0.65\pm0.05$} \\ \rowcolor{green!40}
        HsO-VP Tran. & $84.7\pm1.4$ & $73.5\pm2.4$ & $96.6\pm2.2$ & $85.6\pm2.8$ & $0.67\pm0.04$ \\ \midrule[0.6pt]
        Expert~\cite{roach} & $85.5\pm2.6$ & $75.0\pm4.1$ & $100.0\pm0.0$ & $86.4\pm3.9$ & $0.67\pm0.02$ \\
    \bottomrule[1.0pt]
    \end{tabular}}\vspace{-20pt}
\end{table}
\begin{figure*}[tb!]
    \centering\vspace{-0.1cm}
    \subfigure[\footnotesize OPAL Skill Extraction]{
    \raisebox{-0.14cm}{\label{fig:opal_tsne}\includegraphics[width=0.3\textwidth,height=3.6cm]{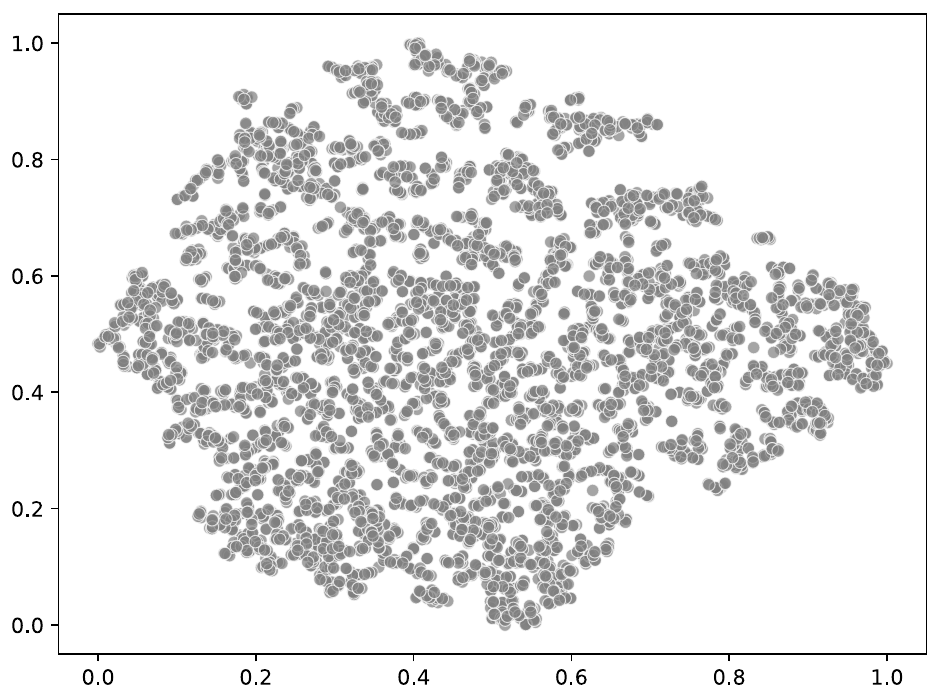}}
    }\hspace{0.1cm}
    \subfigure[\footnotesize Skill Extraction]{
    \raisebox{-0.12cm}{\label{fig:hsovp_tsne}\includegraphics[width=0.3\textwidth,height=3.63cm]{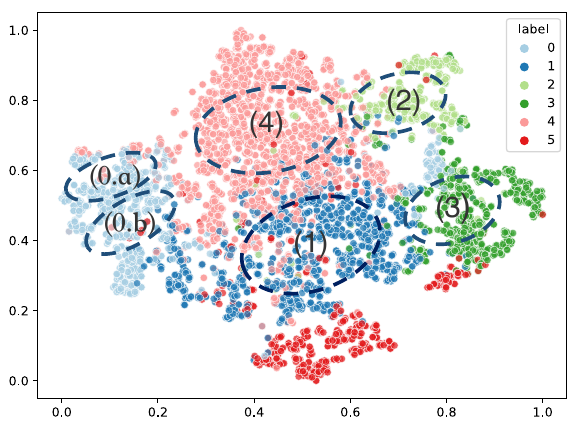}}
    }
    \subfigure[\footnotesize Skill Semantics]{
    \raisebox{-0cm}{\label{fig:skill_vis}\includegraphics[width=0.32\textwidth,height=3.48cm]{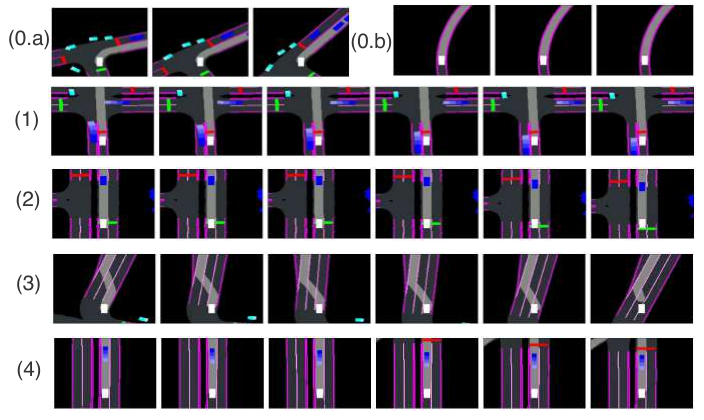}}
    }
    \vspace{-10pt}
    \captionsetup{font={footnotesize}}
    \caption{Visualization of learned skills. (a): T-SNE~\cite{tsne} visualization of skill embeddings learned by OPAL~\cite{opal}. (b): T-SNE visualization of skill embeddings $z$ (i.e. continuous skills) output by HsO-VP's sequence encoder, where the color labels are determined by $y$ (i.e. discrete skills) obtained after the gumbel-softmax operation~\cite{gumbel}. (c): Visualization of the skill sequences extracted from the corresponding areas in the left image.}\vspace{-15pt}
    \label{fig:skill_int}
\end{figure*}
First, we implement baselines and HsO-VP, and test them at training (Town03) and new (Town05) driving scenarios (\highlight{\textbf{Q1}}). Results are recorded in Tab.~\ref{tab:tt_carla} and Tab.~\ref{tab:nn_carla}. Notably, planners achieve higher driving scores at the new scenario because Town03 is inherently more complex than Town05~\cite{carla}. 

Analyzing results in the tables, it becomes evident that offline RL algorithms (CQL~\cite{cql} and IQL~\cite{iql}) outperform IL algorithms (BC and MARWIL~\cite{marwil}) at both training and new scenarios, in line with our claims in the preceding texts. 
Furthermore, we can observe that hierarchical offline RL algorithms (IRIS~\cite{iris} and OPAL~\cite{opal}) can indeed better address long-horizon planning tasks, achieving performance improvement compared to CQL and IQL. However, upon closer examination, OPAL only exhibits limited improvement in driving scores compared to the vanilla IQL (in absolute value, 1.4\% improvement at training scenarios and 0.9\% at new scenarios), which implies that OPAL suffers from posterior collapse~\cite{oversmoothvae,oversmoothvae2} and fails to extract highly effective skills, better shown in Fig.~\ref{fig:opal_tsne}.

In contrast, when we reformulate the generation process and introduce additional discrete skill variables, the trained HsO-VP planner achieves optimal results across metrics at both training and new scenarios. For the most significant driving score metric, it outperforms the strongest baseline IRIS by about 4.0\% in absolute value at training scenarios and 6.4\% at new scenarios. The significantly higher infraction score and normalized reward indicate that HsO-VP obtains higher driving efficiency and safety simultaneously. These results suggest that the hierarchical skills extracted by HsO-VP enable long-term reasoning for offline RL algorithms.
\vspace{-20pt}\subsection{Ablation Study}
The key component of HsO-VP is its two-branch VAE structure. If a vanilla VAE is used instead, the algorithm just degenerates into OPAL~\cite{opal}, which has been examined in Sec.~\ref{sec:drive_exp} to be significantly weaker than HsO-VP. In this section, we primarily conduct ablations on the training process of HsO-VP (\highlight{\textbf{Q2}}): First, the variant that directly uses extracted skills for policy initialization and tests without downstream training, is denoted as `HsO-VP Raw'; Second, the variant that only trains the high-level policy using offline RL but does not finetune the low-level policy by BC, is called `HsO-VP w/o BC'. The results are recorded in Tab.~\ref{tab:tt_carla} and Tab.~\ref{tab:nn_carla}.

The ablation results experiments show pronounced differences. We find that both discarding the finetuning of low-level policies and offline RL of high-level policies lead to performance losses across all metrics. The impact of not performing low-level finetuning is relatively small, indicating that the latent variables in the skill extraction phase are already well aligned with the execution of low-level actions. However, `HsO-VP Raw' only obtains limited performance improvement compared to IL and is inferior to vanilla offline RL algorithms. Essentially, the skill extraction process does not take into account the reward signals but simply reconstructs expert actions, albeit considering actions over multiple timesteps. Therefore, the skill-based offline RL process is crucial for the performance of HsO-VP.

\vspace{-5pt}\subsection{Skill Analysis}
In this section, we explore the properties of extracted skills from HsO-VP through further experiments (\highlight{\textbf{Q3}}).

\textbf{Interpretability.} In Fig.~\ref{fig:hsovp_tsne}, we visualize the final output $z$ from the two-branch sequence encoder using t-SNE~\cite{tsne}. We analyze the role of the discrete layer by using latent variables $y$ obtained from the discrete branch as labels. From the figure, we can observe that sequences with the same color are mostly clustered together, indicating that the gumbel-softmax operation~\cite{gumbel} assigns skills from different discrete options to different networks in the ensemble, thereby amplifying the differences between skill embeddings $z$. As a result, we can observe that the extracted skills of HsO-VP are obviously more distinguishable than OPAL~\cite{opal} in Fig.~\ref{fig:opal_tsne}. In Fig.~\ref{fig:skill_vis}, we visualize representative sequences from selected clusters, it can be seen that different colored blocks correspond to distinct discrete skill choices (from 0 to 4, right turn, stop, accelerate, left turn, and go straight, respectively). Furthermore, the clusters 0.a and 0.b in the same color (i.e. within the same skill option) stand for sharp and mild right turns, respectively, indicating flexible execution styles for actions in a discrete skill. These results demonstrate that HsO-VP captures both the discrete options and continuous variations of skills, providing evidence for the interpretability of skills.

\textbf{Transferability~\cite{skillprior}.} In Sec.~\ref{sec:drive_exp}, we have deployed HsO-VP at new driving scenarios and validated its strong performance. Here we study a different setting: \emph{leverage the skill extraction module obtained from training scenarios to label skills at new (testing) scenarios, then use them instead of original training data for downstream training.} Essentially, we are investigating the transferability of learned skills, analyzing whether they can be directly used to facilitate planning in unseen driving tasks. The results are also recorded in Tab.~\ref{tab:nn_carla}, highlighted in green and labeled as `Tran.'. 
Notably, `OPAL Tran.' which extracts skills through a vanilla VAE only obtains limited performance improvements compared to OPAL (0.7\% driving score difference), which implies that the learned skills from training data do not align with data from new driving tasks. In contrast, `HsO-VP Tran.' shows significant improvements across all metrics compared to `HsO-VP'. These results prove that HsO-VP succeeds in learning skills that can be transferred to assist planning in new tasks.

\section{Conclusion and Outlooks}
In this paper, we propose HsO-VP to boost offline RL for long-horizon vehicle planning. Specifically, we filter offline driving data and design a two-branch VAE to extract hierarchical skills that capture both discrete options and continuous variations, overcoming posterior collapse. Based on this, we train a high-level policy that outputs skills instead of per-step actions, which serve as temporal abstractions to enable long-term reasoning into the future. Comprehensive experimental results on CARLA prove that HsO-VP extracts interpretable driving skills while consistently outperforming strong baselines at both seen and unseen driving scenarios.

As a pioneering work to leverage skill-based offline RL for vehicle planning, we believe that HsO-VP will be a promising and inspiring framework to enable practical autonomous driving. And there are quite a few interesting follow-up directions, including but not limited to learning driving skills from human-annotated skill data, a more refined generative process that can capture transitions between skills, extracting variable-length driving skills, etc.



\newpage
\bibliographystyle{unsrt}
\bibliography{egbib}

\begin{thebibliography}{10}

\bibitem{carla}
Alexey Dosovitskiy, German Ros, Felipe Codevilla, Antonio Lopez, and Vladlen
  Koltun.
\newblock Carla: An open urban driving simulator.
\newblock In {\em Conference on robot learning}, pages 1--16. PMLR, 2017.

\bibitem{intersim}
Qiao Sun, Xin Huang, Brian~C Williams, and Hang Zhao.
\newblock Intersim: Interactive traffic simulation via explicit relation
  modeling.
\newblock In {\em 2022 IEEE/RSJ International Conference on Intelligent Robots
  and Systems (IROS)}, pages 11416--11423. IEEE, 2022.

\bibitem{nuplan}
Holger Caesar, Juraj Kabzan, Kok~Seang Tan, Whye~Kit Fong, Eric Wolff, Alex
  Lang, Luke Fletcher, Oscar Beijbom, and Sammy Omari.
\newblock nuplan: A closed-loop ml-based planning benchmark for autonomous
  vehicles.
\newblock {\em arXiv preprint arXiv:2106.11810}, 2021.

\bibitem{waymo}
Scott Ettinger, Shuyang Cheng, Benjamin Caine, Chenxi Liu, Hang Zhao, Sabeek
  Pradhan, Yuning Chai, Ben Sapp, Charles~R Qi, Yin Zhou, et~al.
\newblock Large scale interactive motion forecasting for autonomous driving:
  The waymo open motion dataset.
\newblock In {\em Proceedings of the IEEE/CVF International Conference on
  Computer Vision}, pages 9710--9719, 2021.

\bibitem{ilreduction}
St{\'e}phane Ross, Geoffrey Gordon, and Drew Bagnell.
\newblock A reduction of imitation learning and structured prediction to
  no-regret online learning.
\newblock In {\em Proceedings of the fourteenth international conference on
  artificial intelligence and statistics}, pages 627--635. JMLR Workshop and
  Conference Proceedings, 2011.

\bibitem{ilaggregation}
Aditya Prakash, Aseem Behl, Eshed Ohn-Bar, Kashyap Chitta, and Andreas Geiger.
\newblock Exploring data aggregation in policy learning for vision-based urban
  autonomous driving.
\newblock In {\em Proceedings of the IEEE/CVF Conference on Computer Vision and
  Pattern Recognition}, pages 11763--11773, 2020.

\bibitem{offlinerl}
Sergey Levine, Aviral Kumar, George Tucker, and Justin Fu.
\newblock Offline reinforcement learning: Tutorial, review, and perspectives on
  open problems.
\newblock {\em arXiv preprint arXiv:2005.01643}, 2020.

\bibitem{bcql}
Scott Fujimoto, David Meger, and Doina Precup.
\newblock Off-policy deep reinforcement learning without exploration.
\newblock In {\em International conference on machine learning}, pages
  2052--2062. PMLR, 2019.

\bibitem{morel}
Rahul Kidambi, Aravind Rajeswaran, Praneeth Netrapalli, and Thorsten Joachims.
\newblock Morel: Model-based offline reinforcement learning.
\newblock {\em Advances in neural information processing systems},
  33:21810--21823, 2020.

\bibitem{offlinead}
Tianyu Shi, Dong Chen, Kaian Chen, and Zhaojian Li.
\newblock Offline reinforcement learning for autonomous driving with safety and
  exploration enhancement.
\newblock {\em arXiv preprint arXiv:2110.07067}, 2021.

\bibitem{umbrella}
Christopher Diehl, Timo Sievernich, Martin Kr{\"u}ger, Frank Hoffmann, and
  Torsten Bertran.
\newblock Umbrella: Uncertainty-aware model-based offline reinforcement
  learning leveraging planning.
\newblock {\em arXiv preprint arXiv:2111.11097}, 2021.

\bibitem{higorl}
Jinning Li, Chen Tang, Masayoshi Tomizuka, and Wei Zhan.
\newblock Hierarchical planning through goal-conditioned offline reinforcement
  learning.
\newblock {\em arXiv preprint arXiv:2205.11790}, 2022.

\bibitem{drivingskill}
Tong Zhou, Letian Wang, Ruobing Chen, Wenshuo Wang, and Yu~Liu.
\newblock Accelerating reinforcement learning for autonomous driving using
  task-agnostic and ego-centric motion skills.
\newblock {\em arXiv preprint arXiv:2209.12072}, 2022.

\bibitem{hindsight}
Marcin Andrychowicz, Filip Wolski, Alex Ray, Jonas Schneider, Rachel Fong,
  Peter Welinder, Bob McGrew, Josh Tobin, OpenAI Pieter~Abbeel, and Wojciech
  Zaremba.
\newblock Hindsight experience replay.
\newblock {\em Advances in neural information processing systems}, 30, 2017.

\bibitem{sparse}
Elliot Chane-Sane, Cordelia Schmid, and Ivan Laptev.
\newblock Goal-conditioned reinforcement learning with imagined subgoals.
\newblock In {\em International Conference on Machine Learning}, pages
  1430--1440. PMLR, 2021.

\bibitem{hiil}
Eli Bronstein, Mark Palatucci, Dominik Notz, Brandyn White, Alex Kuefler, Yiren
  Lu, Supratik Paul, Payam Nikdel, Paul Mougin, Hongge Chen, et~al.
\newblock Hierarchical model-based imitation learning for planning in
  autonomous driving.
\newblock {\em arXiv preprint arXiv:2210.09539}, 2022.

\bibitem{behavior}
Jian Jing, Elizabeth~C Cropper, Itay Hurwitz, and Klaudiusz~R Weiss.
\newblock The construction of movement with behavior-specific and
  behavior-independent modules.
\newblock {\em Journal of Neuroscience}, 24(28):6315--6325, 2004.

\bibitem{drivingskill1}
Letian Wang, Yeping Hu, Liting Sun, Wei Zhan, Masayoshi Tomizuka, and Changliu
  Liu.
\newblock Transferable and adaptable driving behavior prediction.
\newblock {\em arXiv preprint arXiv:2202.05140}, 2022.

\bibitem{temporalabs}
Richard~S Sutton, Doina Precup, and Satinder Singh.
\newblock Between mdps and semi-mdps: A framework for temporal abstraction in
  reinforcement learning.
\newblock {\em Artificial intelligence}, 112(1-2):181--211, 1999.

\bibitem{rethinking}
Chris Zhang, Runsheng Guo, Wenyuan Zeng, Yuwen Xiong, Binbin Dai, Rui Hu,
  Mengye Ren, and Raquel Urtasun.
\newblock Rethinking closed-loop training for autonomous driving.
\newblock In {\em European Conference on Computer Vision}, pages 264--282.
  Springer, 2022.

\bibitem{skillprior}
Karl Pertsch, Youngwoon Lee, and Joseph Lim.
\newblock Accelerating reinforcement learning with learned skill priors.
\newblock In {\em Conference on robot learning}, pages 188--204. PMLR, 2021.

\bibitem{lmp}
Corey Lynch, Mohi Khansari, Ted Xiao, Vikash Kumar, Jonathan Tompson, Sergey
  Levine, and Pierre Sermanet.
\newblock Learning latent plans from play.
\newblock In {\em Conference on robot learning}, pages 1113--1132. PMLR, 2020.

\bibitem{opal}
Anurag Ajay, Aviral Kumar, Pulkit Agrawal, Sergey Levine, and Ofir Nachum.
\newblock Opal: Offline primitive discovery for accelerating offline
  reinforcement learning.
\newblock In {\em International Conference on Learning Representations}, 2021.

\bibitem{gru}
Kyunghyun Cho, Bart Van~Merri{\"e}nboer, Dzmitry Bahdanau, and Yoshua Bengio.
\newblock On the properties of neural machine translation: Encoder-decoder
  approaches.
\newblock {\em arXiv preprint arXiv:1409.1259}, 2014.

\bibitem{variational}
Cheng Zhang, Judith B{\"u}tepage, Hedvig Kjellstr{\"o}m, and Stephan Mandt.
\newblock Advances in variational inference.
\newblock {\em IEEE transactions on pattern analysis and machine intelligence},
  41(8):2008--2026, 2018.

\bibitem{oversmoothvae}
Yuhuai Wu, Yuri Burda, Ruslan Salakhutdinov, and Roger Grosse.
\newblock On the quantitative analysis of decoder-based generative models.
\newblock In {\em International Conference on Learning Representations}, 2017.

\bibitem{oversmoothvae2}
Christopher~P Burgess, Irina Higgins, Arka Pal, Loic Matthey, Nick Watters,
  Guillaume Desjardins, and Alexander Lerchner.
\newblock Understanding disentangling in $beta$-vae.
\newblock {\em arXiv preprint arXiv:1804.03599}, 2018.

\bibitem{nvae}
Arash Vahdat and Jan Kautz.
\newblock Nvae: A deep hierarchical variational autoencoder.
\newblock {\em Advances in neural information processing systems},
  33:19667--19679, 2020.

\bibitem{gumbel}
Eric Jang, Shixiang Gu, and Ben Poole.
\newblock Categorical reparameterization with gumbel-softmax.
\newblock {\em arXiv preprint arXiv:1611.01144}, 2016.

\bibitem{cql}
Aviral Kumar, Aurick Zhou, George Tucker, and Sergey Levine.
\newblock Conservative q-learning for offline reinforcement learning.
\newblock {\em Advances in Neural Information Processing Systems},
  33:1179--1191, 2020.

\bibitem{iql}
Ilya Kostrikov, Ashvin Nair, and Sergey Levine.
\newblock Offline reinforcement learning with implicit q-learning.
\newblock {\em arXiv preprint arXiv:2110.06169}, 2021.

\bibitem{chauffeurnet}
Mayank Bansal, Alex Krizhevsky, and Abhijit Ogale.
\newblock Chauffeurnet: Learning to drive by imitating the best and
  synthesizing the worst.
\newblock {\em arXiv preprint arXiv:1812.03079}, 2018.

\bibitem{nsm}
Sebastian Starke, He~Zhang, Taku Komura, and Jun Saito.
\newblock Neural state machine for character-scene interactions.
\newblock {\em ACM Trans. Graph.}, 38(6):209--1, 2019.

\bibitem{roach}
Zhejun Zhang, Alexander Liniger, Dengxin Dai, Fisher Yu, and Luc Van~Gool.
\newblock End-to-end urban driving by imitating a reinforcement learning coach.
\newblock In {\em Proceedings of the IEEE/CVF International Conference on
  Computer Vision}, pages 15222--15232, 2021.

\bibitem{mbil}
Anthony Hu, Gianluca Corrado, Nicolas Griffiths, Zachary Murez, Corina Gurau,
  Hudson Yeo, Alex Kendall, Roberto Cipolla, and Jamie Shotton.
\newblock Model-based imitation learning for urban driving.
\newblock In {\em Advances in Neural Information Processing Systems}, 2022.

\bibitem{optimize}
Julius Ziegler, Philipp Bender, Thao Dang, and Christoph Stiller.
\newblock Trajectory planning for bertha—a local, continuous method.
\newblock In {\em 2014 IEEE intelligent vehicles symposium proceedings}, pages
  450--457. IEEE, 2014.

\bibitem{cilqr}
Jianyu Chen, Wei Zhan, and Masayoshi Tomizuka.
\newblock Autonomous driving motion planning with constrained iterative lqr.
\newblock {\em IEEE Transactions on Intelligent Vehicles}, 4(2):244--254, 2019.

\bibitem{minimal}
Scott Fujimoto and Shixiang~Shane Gu.
\newblock A minimalist approach to offline reinforcement learning.
\newblock {\em Advances in neural information processing systems},
  34:20132--20145, 2021.

\bibitem{uncertainty}
Yue Wu, Shuangfei Zhai, Nitish Srivastava, Joshua~M Susskind, Jian Zhang,
  Ruslan Salakhutdinov, and Hanlin Goh.
\newblock Uncertainty weighted actor-critic for offline reinforcement learning.
\newblock In {\em International Conference on Machine Learning}, pages
  11319--11328. PMLR, 2021.

\bibitem{iris}
Ajay Mandlekar, Fabio Ramos, Byron Boots, Silvio Savarese, Li~Fei-Fei, Animesh
  Garg, and Dieter Fox.
\newblock Iris: Implicit reinforcement without interaction at scale for
  learning control from offline robot manipulation data.
\newblock In {\em 2020 IEEE International Conference on Robotics and Automation
  (ICRA)}, pages 4414--4420. IEEE, 2020.

\bibitem{skillmetarl}
Taewook Nam, Shao-Hua Sun, Karl Pertsch, Sung~Ju Hwang, and Joseph~J Lim.
\newblock Skill-based meta-reinforcement learning.
\newblock In {\em International Conference on Learning Representations}, 2022.

\bibitem{tacorl}
Erick Rosete-Beas, Oier Mees, Gabriel Kalweit, Joschka Boedecker, and Wolfram
  Burgard.
\newblock Latent plans for task-agnostic offline reinforcement learning.
\newblock In {\em 6th Annual Conference on Robot Learning}, 2022.

\bibitem{drivingskill2}
Nachiket Deo, Akshay Rangesh, and Mohan~M Trivedi.
\newblock How would surround vehicles move? a unified framework for maneuver
  classification and motion prediction.
\newblock {\em IEEE Transactions on Intelligent Vehicles}, 3(2):129--140, 2018.

\bibitem{mdp}
Martin~L Puterman.
\newblock {\em Markov decision processes: discrete stochastic dynamic
  programming}.
\newblock John Wiley \& Sons, 2014.

\bibitem{kmeans}
Mohiuddin Ahmed, Raihan Seraj, and Syed Mohammed~Shamsul Islam.
\newblock The k-means algorithm: A comprehensive survey and performance
  evaluation.
\newblock {\em Electronics}, 9(8):1295, 2020.

\bibitem{betavae}
Irina Higgins, Loic Matthey, Arka Pal, Christopher Burgess, Xavier Glorot,
  Matthew Botvinick, Shakir Mohamed, and Alexander Lerchner.
\newblock beta-vae: Learning basic visual concepts with a constrained
  variational framework.
\newblock In {\em International conference on learning representations}, 2017.

\bibitem{leaderboard}
Carla team.
\newblock Carla autonomous driving leaderboard.
\newblock 2020.

\bibitem{marwil}
Qing Wang, Jiechao Xiong, Lei Han, Han Liu, Tong Zhang, et~al.
\newblock Exponentially weighted imitation learning for batched historical
  data.
\newblock {\em Advances in Neural Information Processing Systems}, 31, 2018.

\bibitem{tsne}
Laurens Van~der Maaten and Geoffrey Hinton.
\newblock Visualizing data using t-sne.
\newblock {\em Journal of machine learning research}, 9(11), 2008.

\end{thebibliography}

\end{document}